\pdfoutput=1

\documentclass[11pt]{article}

\usepackage[final]{coling}

\usepackage{times}
\usepackage{latexsym}

\usepackage[T1]{fontenc}

\usepackage[utf8]{inputenc}

\usepackage{microtype}

\usepackage{inconsolata}

\usepackage{graphicx}

\usepackage{booktabs}
\usepackage{amsmath}
\usepackage{amsfonts}
\usepackage{algorithm}
\usepackage{algorithmic}
\usepackage{hyperref}
\usepackage{caption}
\usepackage{subcaption}
\usepackage{float}
\usepackage{comment}

\title{SoftLMs: Efficient Adaptive Low-Rank Approximation of Language Models using Soft-Thresholding Mechanism}

\author{
 \textbf{Priyansh Bhatnagar\textsuperscript{*}}, 
 \textbf{Linfeng Wen},
 \textbf{Mingu Kang\textsuperscript{*}}
\\
\\
  University of California San Diego
\\\textsuperscript{*}\small{\{prbhatnagar, mingu\}@ucsd.edu}
}

\begin{document}
\maketitle
\begin{abstract}
Extensive efforts have been made to boost the performance in the domain of language models by introducing various attention-based transformers. However, the inclusion of linear layers with large dimensions contributes to significant computational and memory overheads. The escalating computational demands of these models necessitate the development of various compression techniques to ensure their deployment on devices, particularly in resource-constrained environments. In this paper, we propose a novel compression methodology that dynamically determines the rank of each layer using a soft thresholding mechanism, which clips the singular values with a small magnitude in a differentiable form. This approach automates the decision-making process to identify the optimal degree of compression for each layer. We have successfully applied the proposed technique to attention-based architectures, including BERT for discriminative tasks and GPT2 and TinyLlama for generative tasks. Additionally, we have validated our method on Mamba, a recently proposed state-space model. Our experiments demonstrate that the proposed technique achieves a speed-up of 1.33$\times$ to 1.72$\times$ in the encoder/ decoder with a 50\% reduction in total parameters.
\end{abstract}

\section{Introduction}

\begin{figure*}[ht]
    \centering
    \begin{subfigure}[b]{0.48\textwidth}
        \centering
        \includegraphics[scale=0.32]{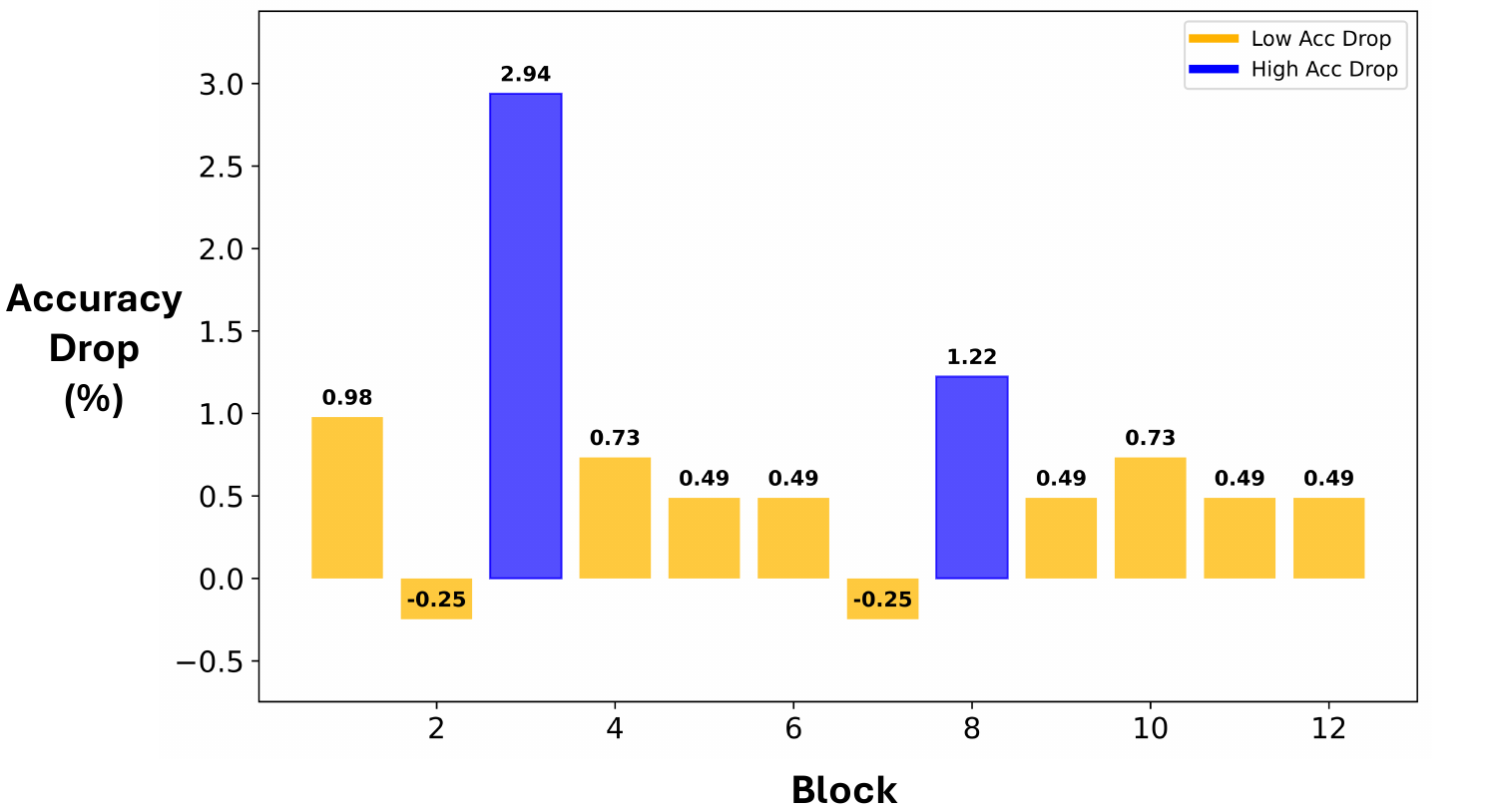}
        \caption{Block-wise accuracy drop }
        \label{fig:a}
    \end{subfigure}
    \hfill
    \begin{subfigure}[b]{0.48\textwidth}
        \centering
        \includegraphics[scale=0.32]{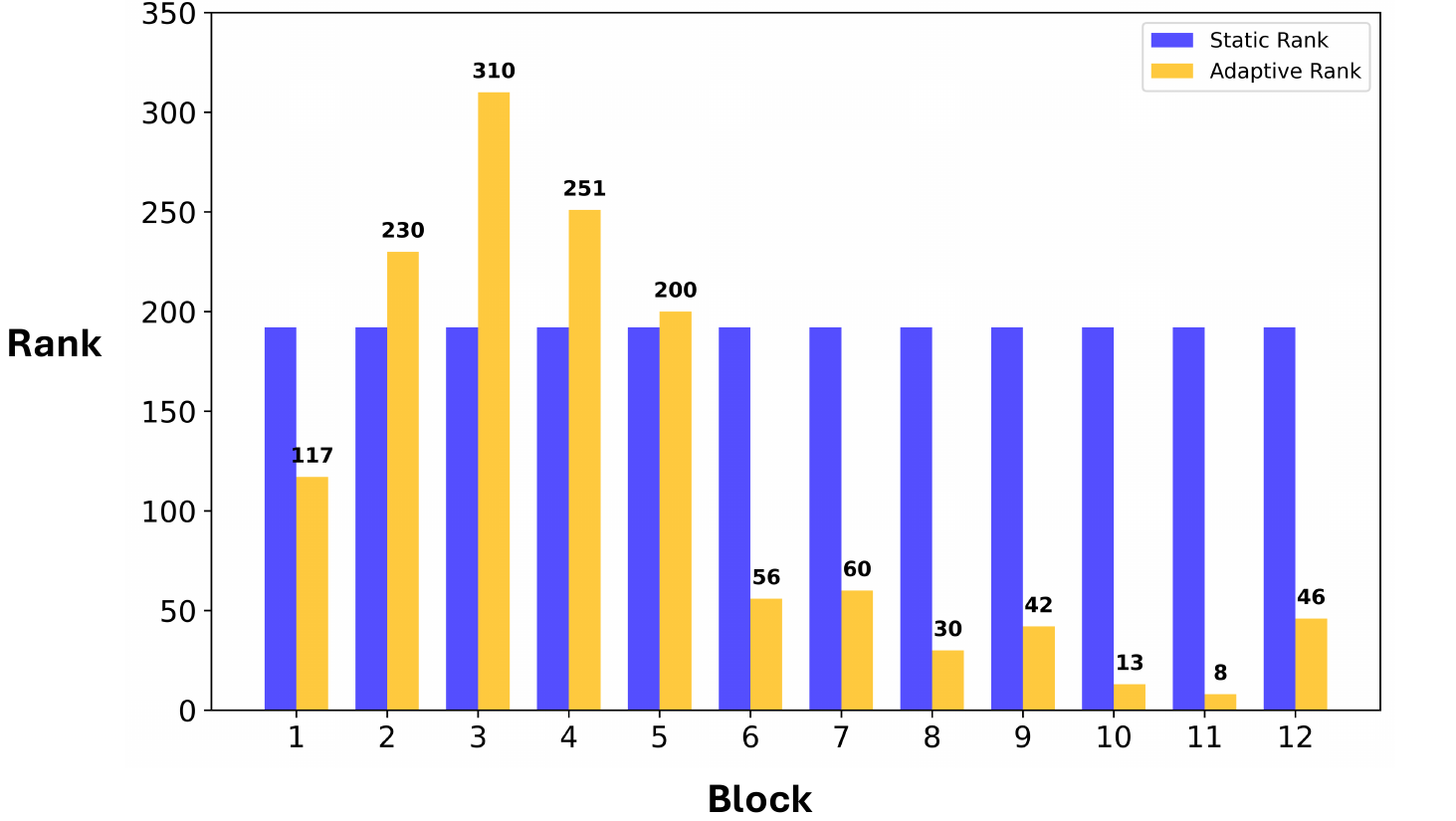}
        \caption{Static and adaptive rank decomposition}
        \label{fig:b}
    \end{subfigure}
    \caption{Accuracy trend of low-rank approximation  for BERT-Base with GLUE MRPC dataset. (a) 
    accuracy drop when individually compressing a BERT block to reduce 50\% of parameters, 
    indicating varying contributions across blocks on overall task performance. (b) Rank of \(W_{Q}\) layer obtained for each block from static vs. proposed adaptive rank decomposition after fine-tuning  for both to have 50\% overall parameter reductions. An F1 score gain of 4.3  is observed from the adaptive rank decomposition (F1: 90.7) over the static method (F1: 86.4).}
    \label{fig: motivation}
\end{figure*}

In the domain of Natural Language Processing (NLP), attention-based models \citep{vaswani2023attention} have achieved unprecedented success across a variety of tasks including language modeling, text classification, and question answering \citep{devlin2018bert, radford2019language}. 
Despite their superior performance, language models (LMs) have been growing increasingly larger, with new models being introduced with higher parameter numbers nearly every day. 
Despite the advancements in powerful GPUs that mitigate some of these overheads, successful deployment on resource-constrained devices remains critical for expanding the applicability of these language models. Specifically, LMs require increasing memory and computational resources as sequence lengths grow longer, posing significant challenges for deployment on edge devices within tight memory and computational  budgets. 

For this reason, model compression has become increasingly important, introducing various techniques including pruning \citep{gordon2020compressing, han2016deep, han2015learning}, knowledge distillation \citep{sun2019patient}, and quantization \citep{zafrir2019q8bert, 8954415}. However, these methods often require extensive computational resources and may not always be feasible for end-users due to their lengthy training times or limited compression effectiveness. An alternative line of research has explored the use of low-rank matrix factorization \citep{golub1971singular} to reduce the parameter count \citep{noach2020compressing, 6638949}. Singular Value Decomposition (SVD) is one of the widely employed  methods to decompose weight matrices into smaller matrices \citep{hsu2022language, schotthöfer2022lowrank}. 
Since one of the decomposed matrices contains singular values ordered by magnitude, which often indicates the importance of each dimension, truncating small singular values effectively compresses the matrix to have lower rank. 
However, this process incurs a new challenge as indiscriminately chosen rank regardless  of their end-to-end impact on the model often degrades the performance significantly. 

Recent studies have proposed modifications to this approach by incorporating task-specific knowledge into the compression process, such as weighting the singular values \citep{hsu2022language, wang2024svdllm}. 
However, these methods assume static importance distributions, meaning that the \textit{low-rank} is set to a constant for all layers or specific groups, which could be sub-optimal in terms of the performance and the parameter count.

On the other hand, our experiments summarized in Figure~\ref{fig: motivation}
 underscore the importance of dynamic rank determination over blocks (layers). Figure~\ref{fig: motivation}(a) demonstrates that each block (layer) of the model contributes differently to overall performance, with specific layers showing higher tolerance for low-rank approximation. Figure~\ref{fig: motivation}(b) also indicates that optimally determining the rank for each block leads to significantly higher performance with the same total parameter count.
Given this motivation, we aim to propose a platform to adaptively find the optimal low-rank to maximize compression with minimal accuracy degradations. Our key objectives of this work are as follows.

\noindent\textbf{Model compression for efficient inference:} 
We aim for efficient inference with a lower number of parameters, allowing deployment in a resource constrained environment. The reduced model complexity also accelerates inference, resulting in lower latency across various language models.

\noindent\textbf{Dynamic and adaptive rank decision:} 
Despite the potential for an optimal combination of ranks for each layer's weight at every processing stage, the prohibitively large search space makes it impractical to find the best combination through exhaustive searching over all possible rank combinations, especially given the recent growth in model sizes. This work automates such a search process during training.

\noindent\textbf{Learnable threshold for singular values:} In the process of truncating the singular values with low magnitude, it is critical to find the optimal threshold to maximize the compression  while minimizing  performance degradation. We employ a single \textit{learnable} threshold parameter for each linear layer to be adapted during the training.  This approach allows each layer to determine the optimal degree of compression based on its specific contribution to the overall task performance. This process simply includes an additional threshold layer while adhering to the standard fine-tuning process based on the pre-trained model, without requiring sophisticated implementation and training methodologies.

Given above motivations, this paper proposes \textit{SoftLM}, a language model that includes a differentiable soft threshold layer, making the threshold learnable and easily deployable in various models. This approach not only achieves competitive performance, with an average accuracy degradation of approximately 1\% compared to the original models, but also provides significant reductions in latency, speeding up inference by upto 1.72× at 50\% reduction in the number of parameters.


\section{Related Work}

\textbf{Static Low-Rank Decomposition}: Recent studies have explored low-rank approximation for deep learning models. Low-Rank Adaptation (LoRA) \citep{hu2021lora} proposes reducing the number of trainable parameters in large language models (LLMs) by factorizing weight updates into low-rank matrices. This approach allows for efficient fine-tuning of pre-trained models without significantly increasing computational resources. However, LoRA primarily focuses on reducing training time without compressing the model used for inference. Similarly, QLoRA \citep{dettmers2023qlora} incorporates quantization techniques to further reduce memory footprint and computational complexity. Another approach, FWSVD \citep{hsu2022language}, introduces a decomposition method that approximates weight matrices using SVD combined with Fisher information, which quantifies the amount of information in the specific matrix. Additionally, ASVD \citep{yuan2024asvd} scales the weight matrix based on the activations' distribution for the layer. However, ASVD suffers from limited parameter reduction of 10\% to 20\% while FWSVD 
suffers from the noticeable performance degradation. 
In SVD-LLM \citep{wang2024svdllm}, the truncation of singular values is directly controlled by introducing a compression loss unlike in FWSVD and ASVD. However, these works only support static rank across all the layers without incorporating the layer-wise impact on performance.

\noindent\textbf{Dynamic low-Rank decomposition}: Rather than employing a static rank for all layers, Low-Rank Adaptation (AdaLoRA) \citep{zhang2023adalora} adjusts the rank based on learning dynamics and task complexity, thereby improving efficiency and effectiveness in minimizing the training complexity, but not for the inference.
For the inference tasks, RankDyna \citep{hua-etal-2023-dynamic} proposed an importance-driven dynamic rank adjustment for low-rank decomposition in LLMs, leveraging gradient-based techniques to determine the optimal rank for each layer during training. However, as well noted by the authors, RankDyna suffers from significant memory overhead due to the need to store entire gradients for importance calculations during fine-tuning.

\section{Background}

\begin{figure*}[ht]
  \centering
  \includegraphics[trim=0pt 0pt 0pt 0pt, clip, width=\textwidth]{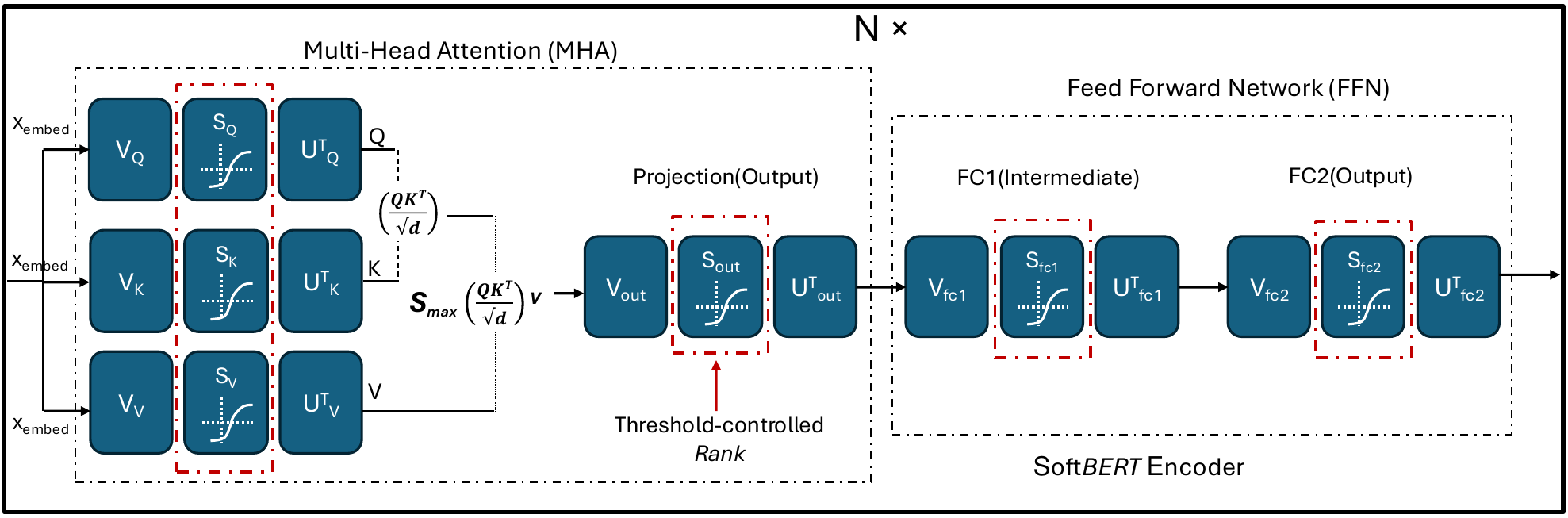}
  \caption{SoftBERT encoder with $N$ blocks, where 
  $W_{Q}$, $W_{K}$, $W_{V}$, $W_{proj}$ in Multi-Head Attention and $W_{fc1}$ and $W_{fc2}$ in the Feed Forward Network are substituted with  U, S  and V. The module S employs the Soft Threshold function to clamp the singular values in $\Sigma$, enabling dynamic rank for each block adaptively in fine-tuning.}
  \label{fig:architecture}
\end{figure*}

\subsection{Low rank approximation}
Singular Value Decomposition (SVD) performs decomposition of a matrix, i.e.,  for any matrix $W\in\mathbb{R}^{M \times N}$, SVD is given by:
\begin{equation}
W = U \Sigma V^T
\end{equation}
where: $U$ is an $M \times M$ orthogonal matrix, $V$ is an $N \times N$ orthogonal matrix, $\Sigma$ is an $M \times N$ diagonal matrix with non-negative real numbers on the diagonal. Elements on the diagonal of $\Sigma$, known as singular values, are denoted by $\sigma_1, \sigma_2, \dots, \sigma_r$ where $\sigma_1 \geq \sigma_2 \geq \cdots \geq \sigma_r > 0$ and $r$ is the rank of matrix $W$. 
To reduce dimensionality, a truncated SVD is commonly used:
\[
W \approx W_k = U_k \Sigma_k V_k^T
\]
where $U_k$, $\Sigma_k$, and $V_k$ contain only the first $k$ columns, singular values, and rows, respectively. 

\subsection{Attention Based Language Models}
Attention-based models \citep{vaswani2023attention} typically comprise of two main modules: Multi Head Attention (MHA) and Feed Forward Network (FFN).  The attention mechanism selectively focuses on different parts of the input sequence when generating an output. 
The fundamental principle is to compute a weighted sum of input values ($V$), where the weights are determined by the similarity between a query ($Q$) and a set of keys ($K$) as follows, where $Q$, $V$ and $K$ are $N\times d_k$-dim matrices with $N$ being the sequence length:
\begin{equation}
    Q = W_Q(x), K = W_K(x), V = W_V(x)
\end{equation}
\begin{equation}
    Attn(Q, K, V) = softmax\left(\frac{QK^T}{\sqrt{d_k}}\right)V
\end{equation}
where \( x \) is the input sequence, $\sqrt{d_k}$ and \( W_Q \), \( W_K \), and \( W_V \) are learnable weight matrices for the generation of queries, keys, and values, respectively.
The attention scores are computed using the dot product of the query with each key stored in each column of $K$, followed by a softmax function to obtain the weights.
The attention module is typically followed by an additional linear layer .
Feed Forward Network (FFN) consists of two linear layers ($W_{fc1}$ and $W_{fc2}$). The first layer expands the dimension of the input, while the second layer reduces it back to the original dimension.





\subsection{SSMs and Mamba}

State-space models (SSMs) are mathematical models used to describe the behavior of dynamic systems. A general state-space model is defined by:

\begin{equation}
\begin{aligned}
    h_{t} &= A h_{t-1} + B x_{t-1}\\
    y_t &= C h_t 
\end{aligned}
\end{equation}
where $h_t$, $x_t$, and $y_t$  are the state, input, and output vectors or matrices at time step $t$. The $A$, $B$ and $C$ are matrices that define the system dynamics. 

Mamba is a selective state-space based model which includes above state-space component surrounded by two projection layers  at the start ($in_{proj}$) and at the end ($out_{proj}$) in each block. 
Unlike the conventional SSM, the $B, C$, and $\Delta$ are made input-dependent by being generated from the linear operation on the input. 
The $in_{proj}$ and $out_{proj}$ layers take the dominant portion (e.g., 94\%) of the parameters in each block whereas the number of parameters inside SSM is 
significantly less. Thus, we focus on these two layers in this  paper.

\section{Methodology}
This section introduces SoftLM based on the proposed SVD compression method with a differentiable thresholding called \textit{Soft Threshold} as shown in Figure~\ref{fig:architecture}, which provides an example on BERT. A detailed algorithm and training methodology are provided to adaptively compress the linear layers, achieving an optimal balance between accuracy and compression.

\subsection{Compressed Linear Layer}
We first replace the standard linear layer with a decomposed linear layer including three modules, $U, \Sigma$ and $V$. Modules for $U \text{ and } V$ share the same structure, which is similar to the vanilla SVD. 
A soft threshold layer $\mathcal{T}h_{s}$ 
(described in Section 4.3) is applied on the $\Sigma$. The conventional linear operation of $y = x \cdot W^T$ is transformed as:
\begin{equation}
 y = x \cdot V \cdot \mathcal{T}h_{s}(\Sigma) \cdot U^T
\end{equation}
where \( x \) is the input, \( W \) is the weight matrix, and \( y \) is the output. 

\subsection{Learnable Threshold on Singular Values}

\begin{figure}[t]
\centering
\includegraphics[scale=0.4]{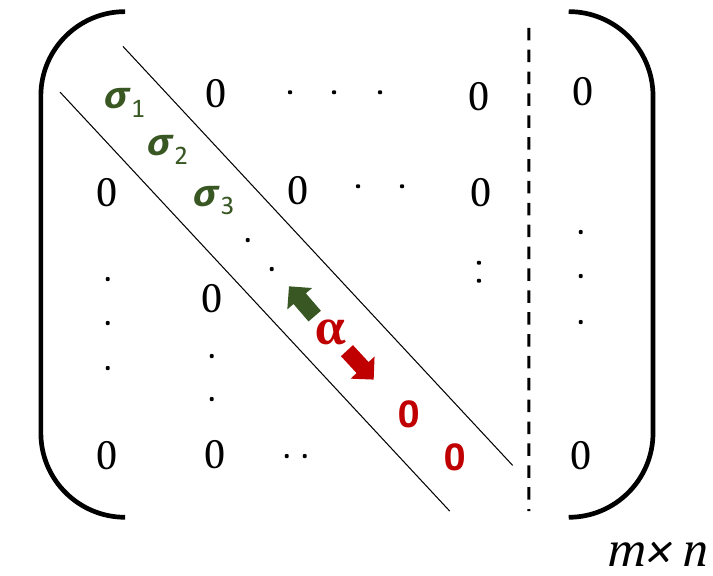}
\caption{Learnable threshold $(\alpha)$ for selecting top singular values in $\Sigma$; values below $(\alpha)$ are set to zero.}
\label{fig: thres}
\end{figure}

The descending order of singular values in the 
$\Sigma$ matrix by magnitude reflects the decreasing importance of the corresponding singular vectors in the SVD matrices (Figure~\ref{fig: thres}). 
 Our technique employs a gradient-based learnable threshold ($\alpha$) to select a limited set of most consequential singular values. This  threshold traverses diagonally across the $\Sigma$ matrix during the training so that values in $\Sigma$ below the threshold are replaced by zero. 
A higher threshold corresponds to a greater level of compression. Consequently, the corresponding rows and columns of the orthogonal matrices $U$ and $V$, respectively, are also neglected. This selective removal of singular vectors allows  effective dimensionality reduction. The threshold is set as a learnable parameter and is distinct for each layer to be adjusted given the varying importance of each linear layer.

\begin{figure}[t]
    \centering
    \begin{subfigure}[b]{0.47\columnwidth}
        \centering
        \includegraphics[scale=0.35]{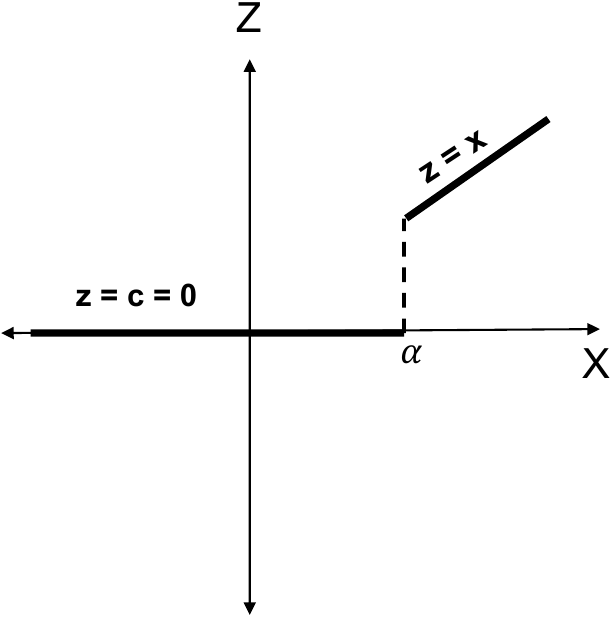}
        \caption{Hard Threshold}
        \label{fig:a}
    \end{subfigure}
    \begin{subfigure}[b]{0.47\columnwidth}
        \centering
        \includegraphics[scale=0.35]{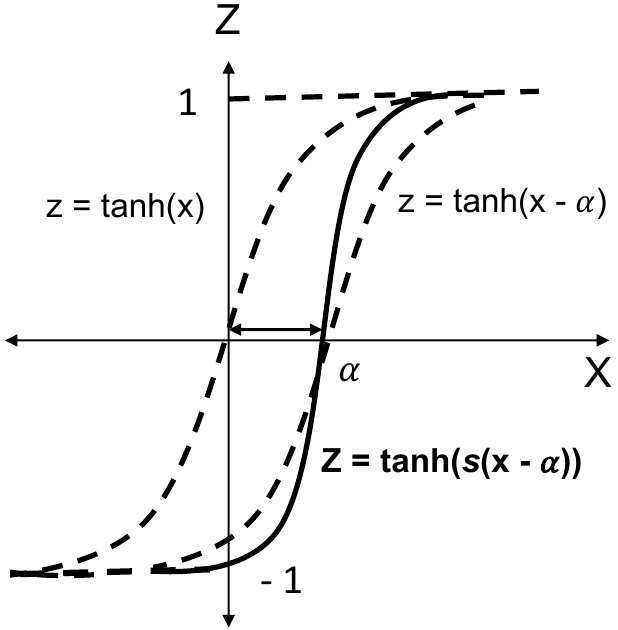}
        \caption{Tanh Function}
        \label{fig:b}
    \end{subfigure}
    
    \vspace{0.3cm}
    
    \begin{subfigure}[b]{0.95\columnwidth}
        \centering
          \includegraphics[scale=0.45]{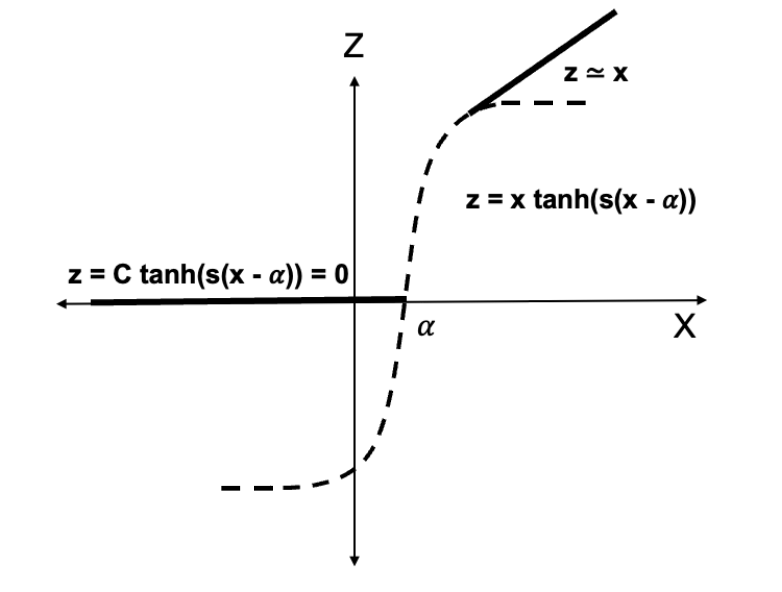} 
        \caption{Soft Threshold}
        \label{fig:b}
    \end{subfigure}
    \caption{Threshold functions. (a) conventional non-differentiable thresholding, (b) shifted Tanh with sharpness control factor $s$, and (c) differentiable soft thresholding combining above functions}
    \label{fig: soft_thres}
\end{figure}

\subsection{Soft Threshold}
It is known that non-differentiable operations hinder the back-propagation process during training, as gradients cannot flow through such operations. However, the threshold function is inherently non-differentiable, which poses challenges in adapting the $\alpha$ during training.
To facilitate the  gradient flow, we employ a soft thresholding (Figure~\ref{fig: soft_thres}) as follows: 
\begin{equation}
\mathcal{T}h_{s}(x) =
\begin{cases} 
      x \tanh(s (x - \alpha)), & \text{if } x \geq \alpha \\
      c \tanh(s (x - \alpha)), & \text{if } x < \alpha 
\end{cases}
\end{equation}
where \(x\) represents the input, \(s\) is a sharpness control parameter, \(\alpha\) is the cut-off value, and constant \(c\) replaces values less than the threshold in the matrix. \(c\) is chosen to be zero in our case.
This function acts as a smooth approximation of the hard thresholding, allowing gradients to propagate through the network during the training process. Specifically, when \(x\) exceeds the cut-off value \(\alpha\), the soft thresholding function yields \(x\) multiplied by the $\textrm{tanh}$ term. Conversely, for \(x\) less than \(\alpha\), the function yields \(c\), effectively replacing values below the threshold to zero.
\begin{table*}[t]
\small
  \centering
  \begin{tabular}{lccccccc}
    \toprule
    \textbf{Model} & \textbf{\#Para} & \textbf{MRPC} & \textbf{COLA} & \textbf{QNLI} & \textbf{QQP} & \textbf{SST2} & \textbf{G-Avg} \\
    \midrule
    BERT-Base (w/o \textit{emb})    & 86M \\
    \hspace{1em} + \textit{embeddings} & 110M & 88.0 & 56.2 & 91.3 & 87.8 & 93.0 & 83.3\\
    \midrule
    DistilBERT   & 67M  & 87.5 & 51.3 & 89.2 & 86.7 & 91.3 & 81.20 \\
    MiniLMv2     & 67M  & 89.1 & 43.3 & \textbf{90.6} & 86.7 & 91.4 & 80.2 \\
    BERT6-KD    & 67M  & 86.2 & -    & 88.3 & -    & 91.5 & - \\
    BERT6-PKD    & 67M  & 85.0 & -    & 89.0 & -    & \textbf{92.0} & - \\
    SVD-BERT     & 67M  & 86.4 & 40.5 & 89.0 & 86.5 & 90.1 & 78.5 \\
    \midrule
    SoftBERT (w/o \textit{emb})    & \hspace{3em} 43M (--50\%)  \\
    \hspace{1em} + \textit{embeddings} & 67M & \textbf{90.7} & \textbf{54.6} & 89.5 & \textbf{87.6} & 90.4 & \textbf{82.6} \\
    \bottomrule
  \end{tabular}
  \caption{A comparison of SoftBERT with prior works in terms of parameter count and performance (reported under the name of datasets). G-Avg: the average performance  for GLUE tasks.}
  \label{tab: bert result}
\end{table*}

\begin{figure}[t]
  \centering
  \includegraphics[scale = 0.5]{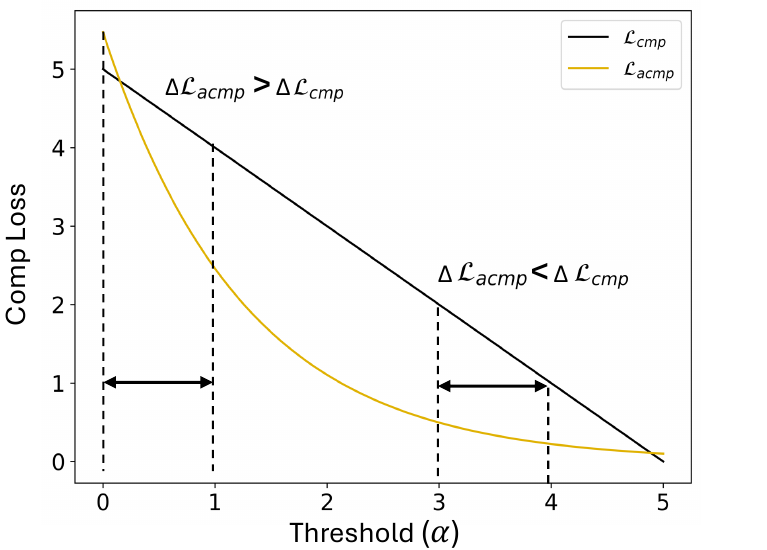}
  \caption{The magnitude of compression loss over the threshold $\alpha$  with a conventional linear loss ($\mathcal{L}_{cmp}$) and the proposed adaptive loss ($\mathcal{L}_{acmp}$).}
  \label{fig: adaptive loss}
\end{figure}

\subsection{Adaptive Loss for Compression}
To maximize the compression and the performance at the same time, we employ the total loss function  defined as:
\begin{equation}
\mathcal{L}_{tot} = \mathcal{L}_{acc} + \gamma \cdot \mathcal{L}_{cmp}
\end{equation}
where $\mathcal{L}_{acc}$ is a cross entropy loss to maximize the performance while $\mathcal{L}_{cmp}$ is the compression loss and $\gamma$ is the balancing factor for regularization. For $\mathcal{L}_{cmp}$, we  simply sum the thresholds as: 
\begin{equation}
\mathcal{L}_{cmp} = -\sum_i \alpha_i
\end{equation}
Here, \(\alpha_i\) is the learnable threshold for the \(i\)-th block. While the \(\alpha_i\) is set to be zero at the beginning of fine-tuning, this loss term gradually increases  the $\alpha_i$  to truncate more singular values. 

However, a linear loss term continuously pressures for more compression at any rank. Ideally, the loss should adjust based on the fine-tuning phase, i.e., a larger loss is preferable at the beginning to accelerate compression whereas the compression should slow down to allow for performance recovery  once the model is sufficiently compressed.  To enable such a capability, 
we introduce an adaptive compression loss ($\mathcal{L}_{acmp}$) defined as follows: 
\begin{equation}
\mathcal{L}_{acmp} = \sum_i e^{-\alpha_i}
\end{equation}
By exploiting the exponential term, the loss reduces to be negligible once the $\alpha_i$ increases enough (Figure~\ref{fig: adaptive loss}). The following total loss is employed in this paper by reflecting the $\mathcal{L}_{acmp}$ as:
\begin{equation}
    \mathcal{L}_{tot} = \mathcal{L}_{acc} + \gamma \cdot \mathcal{L}_{acmp}
\end{equation}
\subsection{Fine-tuning Algorithm}
In this section, we describe the detailed procedure of fine-tuning. It should be noted that the learnable threshold ($\alpha$) is frozen once the total parameter count reaches the target value by changing the learning rate to be zero for the $\alpha$s.
At the end of fine-tuning, we merge $S=\mathcal{T}h_s(\Sigma)$ and $V$ into a matrix to further save memory.

\begin{algorithm}
\caption*{\textbf{Algorithm: }Adaptive Rank using SoftThreshold}
\label{alg:Adaptive Rank Using SoftThreshold}
\begin{algorithmic}
    \REQUIRE $P$: Target parameter number 
    \STATE Initialize $\alpha_i \gets 0$
    \STATE SVD: $U\Sigma V^T \gets W$ 
    \STATE SoftThreshold: $\mathcal{T}h_s(\Sigma) \gets \Sigma$
    \WHILE{$\#\text{param} > P$}
        \STATE Compute $\mathcal{L}_{tot} = \mathcal{L}_{acc} + \gamma \cdot \mathcal{L}_{acmp}$
        \STATE Optimizer step (AdamW)
    \ENDWHILE

    \STATE Freeze threshold $(\alpha_i)$ parameters
    \FOR{$epoch$ in (0, $N_{epoch}$) }
        \STATE Compute $\mathcal{L}_{tot} = \mathcal{L}_{acc} + \gamma \cdot \mathcal{L}_{acmp}$
        \STATE Optimizer step (AdamW)
    \ENDFOR
    \STATE $//$ End-of-fine-tuning
    \STATE Compute $VS = V \cdot \mathcal{T}h_s(\Sigma)$ 
    \STATE Store ($U^{T}, VS$)
    \ENSURE $\#\text{param}$ in ($U$ and $VS$) $<$ $\#\text{param}$ in $W$ 
\end{algorithmic}
\end{algorithm}

\section{Experiments}
\subsection{Models and Datasets}
Extensive experiments are conducted on a diverse set of language models and datasets  to evaluate the efficacy of the proposed method. 
We applied the proposed technique on models: BERT-Base \citep{devlin2018bert}, GPT2-medium \cite{radford2019language}, Mamba-130M \citep{gu2024mamba} and TinyLlamav1.1 \cite{zhang2024tinyllamaopensourcesmalllanguage} to produce SoftBERT, SoftGPT2, SoftMamba and SoftTinyLlama, respectively.  BERT-Base is evaluated on classification tasks such as MRPC, COLA, SST-2, QNLI, and QQP selected from the popular GLUE benchmark \citep{wang2019glue}. While MRPC and QQP were evaluated using the F1-score, COLA was assessed using the Matthews Correlation Coefficient (MCC). SST-2 and QNLI were evaluated based on accuracy. Similarly, Mamba-130M is evaluated on the  GLUE datasets. In Table~\ref{tab: bert result}, the target compression ratio (CR) for SoftBERT is chosen as 0.50 for a fair evaluation and comparison convenience with other references. Table~\ref{tab: bert var result} shows the varying performance of SoftBERT with respect to the compression ratios for the five GLUE datasets. GPT2-medium and TinyLlama are evaluated on WikiText-2 \cite{wikitext2} by measuring the perplexity score. 


\begin{table}[H]
\small
\captionsetup{singlelinecheck=false, skip=5pt}
  \centering
  \begin{tabular}{lccc}
    \toprule
    \textbf{\#Para (encoder)} & \textbf{CR} & \textbf{G-Avg} \\
         \midrule
    86M & 0.00 & 83.1\\ 
    \midrule
    64M & 0.25 & 83.0\\
    43M & 0.50 & 82.6\\
    22M & 0.75 & 78.5\\
    \bottomrule
  \end{tabular}
  \caption{Performance variation (GLUE-Average) of SoftBERT with respect to the compression ratio (CR).}
  \label{tab: bert var result}
\end{table}

\begin{table*}[h]
\small
  \centering
  \begin{tabular}{lcccccc}
    \toprule
    \textbf{Model} & \textbf{\#Para} & \textbf{COLA} & \textbf{SST2} & \textbf{QNLI} & \textbf{QQP} & \textbf{Avg} \\
    \midrule
    S4            & 131M  & 23.0 & 87.0 & 72.14 & 89.6 & 67.9\\
    \midrule
    Mamba (\textit{in/out\_proj})    & 85M & & & & &\\
    \hspace{1em} + \textit{misc} & 130M  & 80.5  & 92.0  & 88.5  & 89.0 & 87.5 \\
    \midrule
    SoftMamba (\textit{in/out\_proj})    & \hspace{3em} 42 M (--50\%)\\
    \hspace{1em} + \textit{misc} & 87M  & 77.3  & 90.3  & 88.1  & 89.6 & 86.3\\
    \bottomrule
  \end{tabular}
  \caption{Task performance (reported under the name of datasets) and parameter count of SoftMamba on GLUE datasets and comparison with S4s and Mamba.}
  \label{tab: mamba_result}
\end{table*}
\subsection{Experiment Results and Comparison with Prior Works}

\begin{figure}[t]
  \centering
  \includegraphics[scale=0.4]{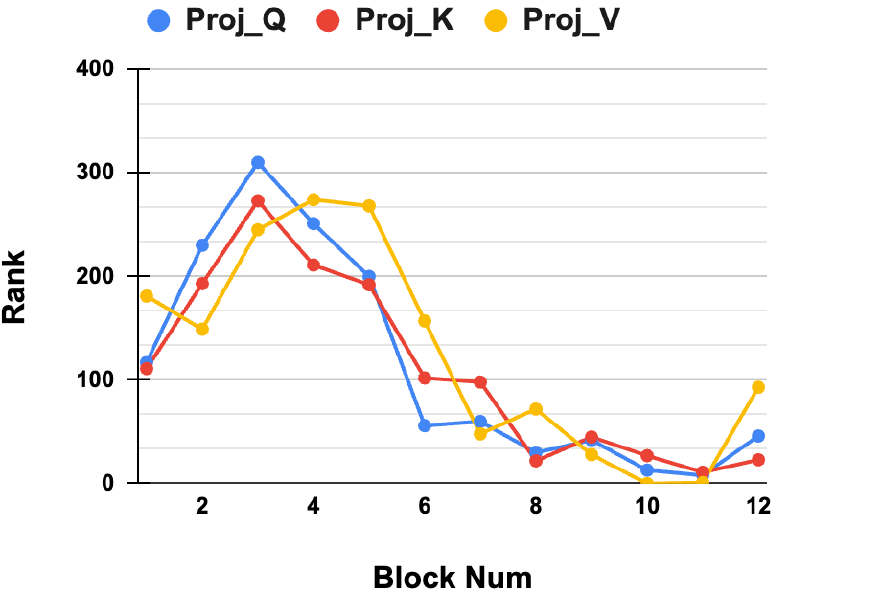}
  \includegraphics[scale=0.4]{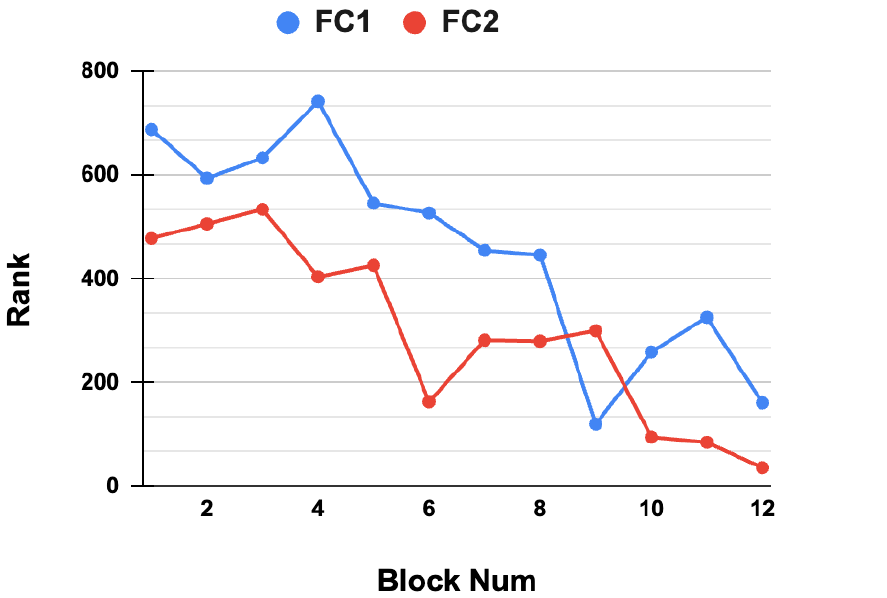}
  \caption{Rank of each block after fine-tuning  SoftBERT on MRPC in MHA (top) and  FFN (bottom).}
  \label{fig:fig7}
\end{figure}

For BERT-Base, the six linear layers ($W_{Q}$, $W_{K}$, $W_{V}$, and $W_{proj}$ in the attention module as well as $W_{fc1}$ and $W_{fc2}$ in the FFN) in each block sum up to 72 layers across the 12 blocks of the entire encoder, accounting for approximately 86 million parameters. The remaining 23 million parameters are allocated for the token, positional, and segment embeddings. In Table~\ref{tab: bert result}, the compression within the encoder is set to CR$=$0.50, reducing the 86 million parameters inside the encoder to 43 million. The parameters for all the embeddings remain untouched. We used the following baselines and benchmarks for comparison with our method: DistilBERT \citep{sanh2020distilbert}, MiniLMv2 \citep{wang2020minilm}, BERT6-KD, PKD \citep{sun2019patient}, and vanilla SVD applied to BERT-Base with fine-tuning. The final model size for each compared model is carefully adjusted to be the same (67M) to ensure a fair comparison. Our method achieves an average performance score of 82.56 in Table~\ref{tab: bert result}, outperforming all the other methods, with a performance drop from the original uncompressed model limited to just 0.7.
It is observed that the rank of each linear layer within the Multi-Head Attention (MHA) and Feed Forward Network (FFN) varies and is learned to balance the performance and the degree of compression (Figure~\ref{fig:fig7}). 
The initial layers demonstrate higher importance and sensitivity to the loss compared to those in the middle and at the end. Consequently, the rank tends to be preserved in the initial layers.

There are four linear layers and a 1-dimensional convolution layer in each block of  Mamba-130m. The number of parameters for 24 blocks amounts to 90M, with approximately 85M contributed by the $in_{proj}$ and $out_{proj}$ layers. Thus, we compress these two layers by 50\%. SoftMamba achieves comparable performance with an average degradation of merely 1.2\% compared to the original Mamba model (Table~\ref{tab: mamba_result}). 

\begin{table}[t]
\small
\captionsetup{singlelinecheck=false, skip=5pt}
  \centering
  \begin{tabular}{lccc}
    \toprule
    \textbf{Model} & \hspace{-1em} \textbf{\#Para(decoder)} & \textbf{wiki-2} & \textbf{CR}\\
         \midrule
    GPT2-medium & 301M  & 16.7 & 0.00\\ 
    \midrule
    SVD-GPT2-medium  & 227M & 23.0 & 0.25\\
    SoftGPT2-medium  & 227M & 18.6 & 0.25\\
    \midrule
    SVD-GPT2-medium  & 151M & 32.1 & 0.50\\
    SoftGPT2-medium  & 151M & 19.6 & 0.50\\
    \midrule
    SVD-GPT2-medium  & 76M & 50.6 & 0.75\\
    SoftGPT2-medium  & 76M & 35.4 & 0.75\\
    \bottomrule
  \end{tabular}
  \caption{Performance  variation of SoftGPT2-medium in perplexity with respect to CR in comparison with static-rank decomposition.}
  \label{tab: gpt result}
\end{table}

Similarly, each block of GPT-2 contains four 1-dimensional convolution layers, accounting for 301M parameters out of the total 345M across 24 blocks. Table~\ref{tab: gpt result} compares the performance of SoftGPT2-medium against the  GPT2-medium compressed with static-SVD, which applies the same compression ration across all blocks. 
\begin{table}[t]
\small
\captionsetup{singlelinecheck=false, skip=5pt}
  \centering
  \begin{tabular}{lccc}
    \toprule
    \textbf{Model} & \textbf{\#Para(decoder)} & \textbf{wiki-2} & \textbf{CR}\\
         \midrule
    TinyLlamav1.1   & 968M  & 8.9 & 0.00\\ 
    \midrule
    SVD-TinyLlama  & 726M & 13.8 & 0.25\\
    SoftTinyLlama  & 726M & 12.0 & 0.25\\
    \midrule
    SVD-TinyLlama  & 484M & 20.4 & 0.50\\
    SoftTinyLlama  & 484M & 15.1 & 0.50\\
    \midrule
    SVD-TinyLlama  & 242M & 39.2 & 0.75\\
    SoftTinyLlama  & 242M & 19.1 & 0.75\\
    \bottomrule
  \end{tabular}
  \caption{Performance variation  of SoftTinyLlama in perplexity with respect to CR in comparison with static-rank decomposition.}
  \label{tab: tinyllama result}
\end{table}
To further extend our analysis on a bigger model with total parameters greater than one billion, we employ TinyLlama to validate our technique in  Table~\ref{tab: tinyllama result}. We observe a significant performance improvement in the aforementioned models using dynamic rank-based compression compared to static rank-based compression, highlighting the importance of adaptive low-rank approximations.

\subsection{Execution Cost Reductions}
In this section, we provide detailed insights about resource utilization by SoftLMs.

\subsubsection{Savings in Memory and Power}
The proposed technique, evaluated at CR$=$0.50, leads to overall savings of 33-44\%, depending on the model. For BERT-base, which initially has a 440 MB weights file, this corresponds to a reduction of 168 MB, assuming all parameters are in float32 format. This technique reduces not only the memory footprint but also the computational complexity, with Multiply-Accumulate (MAC) operations decreasing linearly with the compression ratio.
\begin{figure}[H]
  \centering
  \includegraphics[width=0.9\columnwidth]{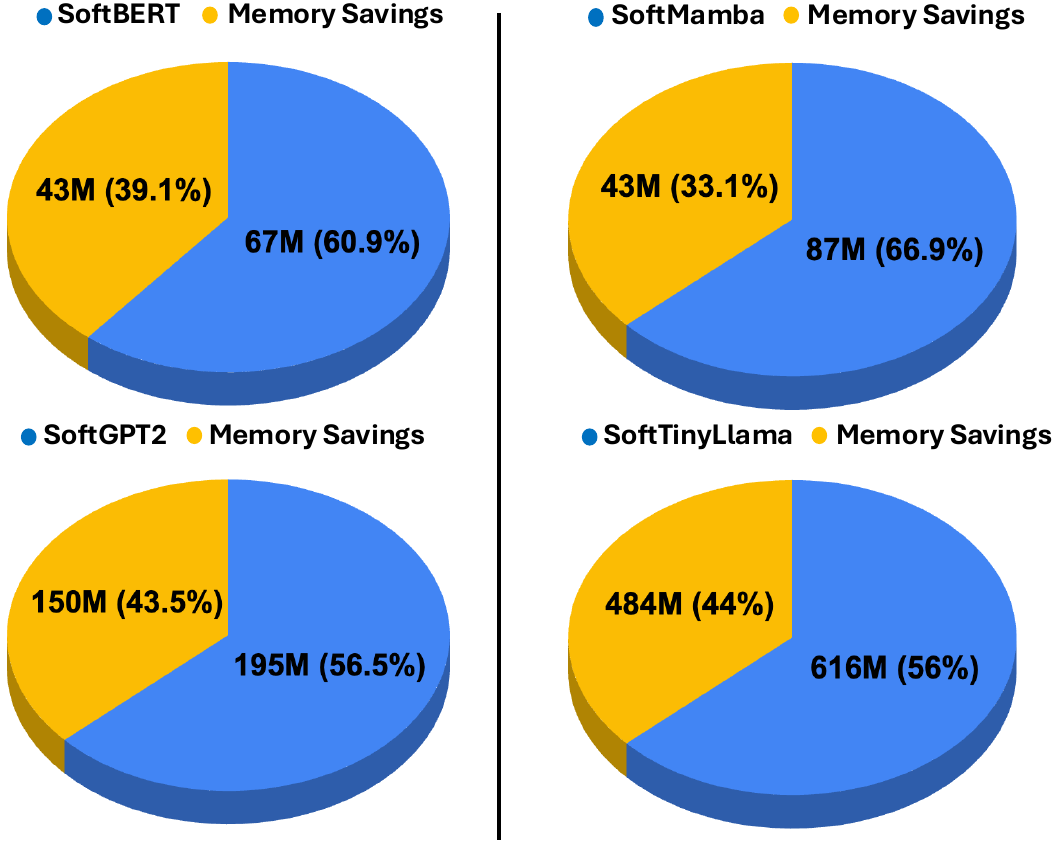}
  \caption{Memory savings in SoftLMs at CR$=$0.50.}
  \label{fig8}
\end{figure}



The power consumption is measured in Watts (W) using the NVIDIA pyNVML library \cite{mlpy} while evaluating the SoftLMs at inference. The reduction in power consumption for BERT and GPT-2 ranges from 30\% to 35\%, while for Mamba and TinyLlama, it varies between 15\% and 20\%.

\subsubsection{Latency Reduction}

\begin{figure}[H]
\centering
\includegraphics[width=0.7\columnwidth]{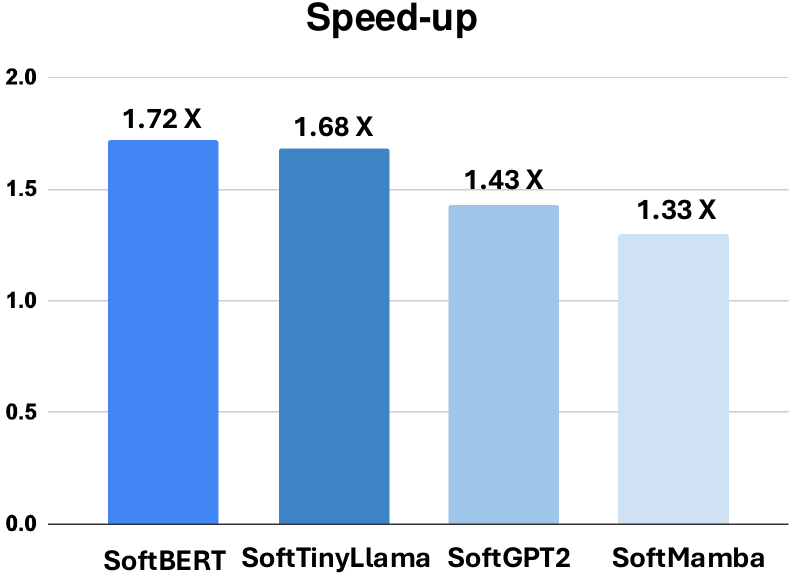}
\caption{Inference speed-up achieved by SoftLMs.}
\label{fig9}
\end{figure}

Latency reduction is measured as the speed-up of the compressed models during inference. Figure~\ref{fig9} shows the average speed-up for SoftBERT, SoftTinyLlama, SoftGPT2, and SoftMamba, each compressed at a CR of 0.50. The speed-up ranges from 1.33$\times$ to 1.72$\times$, with the highest speed-up achieved in BERT, followed by TinyLlama, GPT-2, and Mamba. We attribute the lower speed-up in Mamba to the presence of the SSM block, which is not accelerated by our technique. Additionally, we observe that latency remains similar for a given model, with minimal differences across datasets.

\begin{figure}[H]
  \centering
  \includegraphics[scale=0.65]{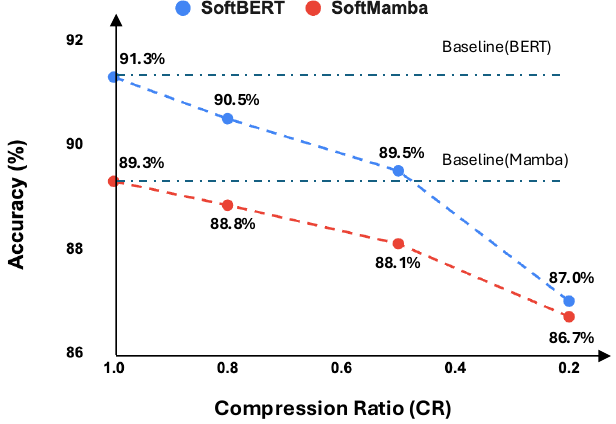}
  \caption{Performance variation with respect to CR in SoftBERT and SoftMamba with QNLI dataset.}
  \label{fig:fig10}
\end{figure}


\subsection{Trade-off in Performance vs. Compression}
Figure~\ref{fig:fig10} illustrates the impact on performance with respect to the  compression ratio for the attention-based BERT and SSM-based Mamba. Although the baseline accuracy of BERT is higher than that of Mamba, the impact of compression is less significant in Mamba, leading to smaller performance degradation at higher compression levels.
For this reason, Mamba is a more feasible model for deployment under stringent resource constraints, particularly due to the absence of Multi-Head Attention (MHA) and its lower performance degradation by the compression.



\section{Conclusion}
We introduce a novel compression method that is easily deployable on any deep learning model with linear layers, adding minimal complexity to the fine-tuning process. By adaptively learning the rank for each layer using a simple thresholding mechanism, our method achieves an optimal balance between task performance and compression. We validate the effectiveness of the proposed technique across various language models, demonstrating that resource-efficient SoftLMs outperform state-of-the-art compression methods.

\section{Limitations}
Although the model is compressed at the end of fine-tuning, the initial decomposition increases its size during the fine-tuning process. This may be less of an issue for smaller models such as BERT and GPT-2, but it presents a challenge when training larger language models. Additionally, while SVD is applied only once during fine-tuning, it is computationally expensive, adding significant complexity to fine-tuning large models. To mitigate this, one could replace the costly SVD with more efficient decomposition methods that begin directly from the truncated rank.


\bibliography{custom}

\appendix

\section{Appendix}
\label{sec:appendix}
\subsection{System Setup}
The system setup that we used includes a NVIDIA GeForce RTX 3090 GPU (total RAM of 24 GB with maximum power limit of 350W) on which experiments for BERT and Mamba were performed. Experiments for TinyLlama and GPT2-medium were performed on a single NVIDIA A100 GPU (total RAM of 80 GB and maximum power limit of 400W). We do not use any distributed or parallel training methods.  

\subsection{Implementation Details}

\subsubsection{Datasets}
The General Language Understanding Evaluation (GLUE) benchmark is a collection of diverse natural language understanding (NLU) tasks designed to evaluate the performance of models on several language understanding challenges. GLUE consists of various tasks including linguistic acceptability, sentence similarity, textual entailment, and natural language inference. 

The benchmark consists of nine tasks, out of which we have evaluated on five. CoLA (Corpus of Linguistic Acceptability) involves determining whether a given sentence is grammatically correct in English. SST-2 (Stanford Sentiment Treebank) requires models to classify the sentiment of a given sentence as either positive or negative. MRPC (Microsoft Research Paraphrase Corpus)  identifies whether two sentences in a pair are semantically equivalent. QQP (Quora Question Pairs) determines whether two questions posted on Quora are paraphrases of each other. QNLI (Question Natural Language Inference) determines whether a given sentence (question) is answerable by another sentence (context). 

The evaluation metric for QNLI and SST2 is accuracy, whereas  it is F1 score for MRPC and QQP, and Matthews Correlation Coefficient (MCC)  for CoLA.

Wikitext-2 is chosen as the generative task which is a set of almost 100 million tokens extracted from Wikipedia's articles. The performance is evaluated through the exponential of validation loss, also commonly referred as the perplexity score.

\subsubsection{Training Setup}
We employed the following hyper-parameters in evaluating BERT and Mamba on GLUE tasks. 
\begin{table}[h]
  \centering
  \begin{tabular}{lc}
    \toprule
    \textbf{Batch Size} & 32\\
    \textbf{Learning Rate (LR)} & {2e-5}\\
    \textbf{LR (threshold $\alpha$)} & {1e-2, 1e-3}\\
    \textbf{Regularizer $(w)$} & {0.001, 0.01, 0.1}\\
    \textbf{Optimizer} & AdamW\\
    \bottomrule
  \end{tabular}
  \caption{Hyper-parameters for fine-tuning.}
  \label{tab:new_models}
\end{table}

For GPT2 and TinyLlama, we used similar settings except the batch size which is set to four for GPT2 and two for TinyLlama due to limited GPU RAM.

\section{SoftLM Rank Distribution}
\begin{figure}[t]
  \centering
  \includegraphics[width=\columnwidth]{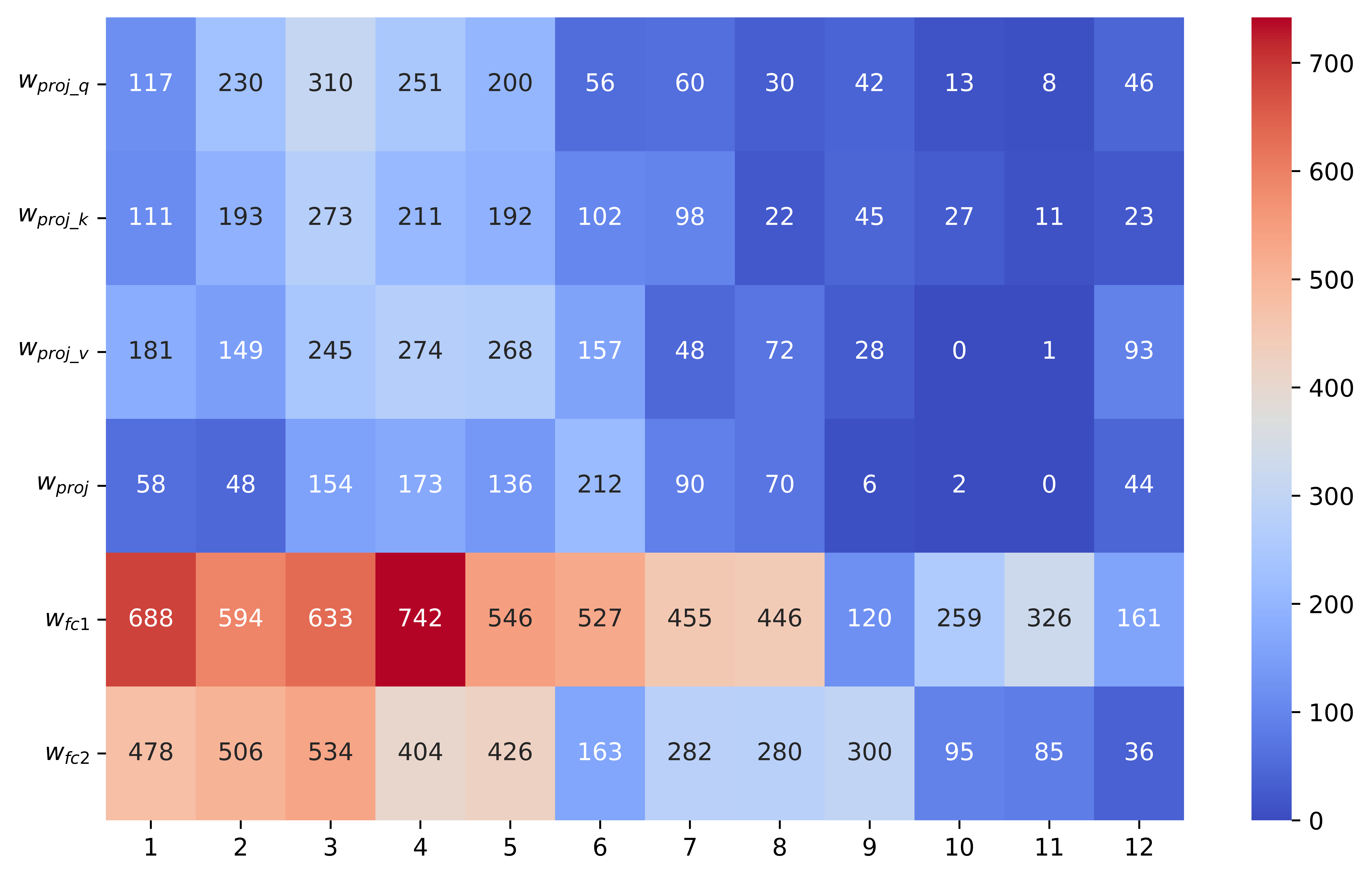}
  \caption{MPRC.}
  \label{fig:mrpc}
\end{figure}

\begin{figure}[t]
  \centering
  \includegraphics[width=\columnwidth]{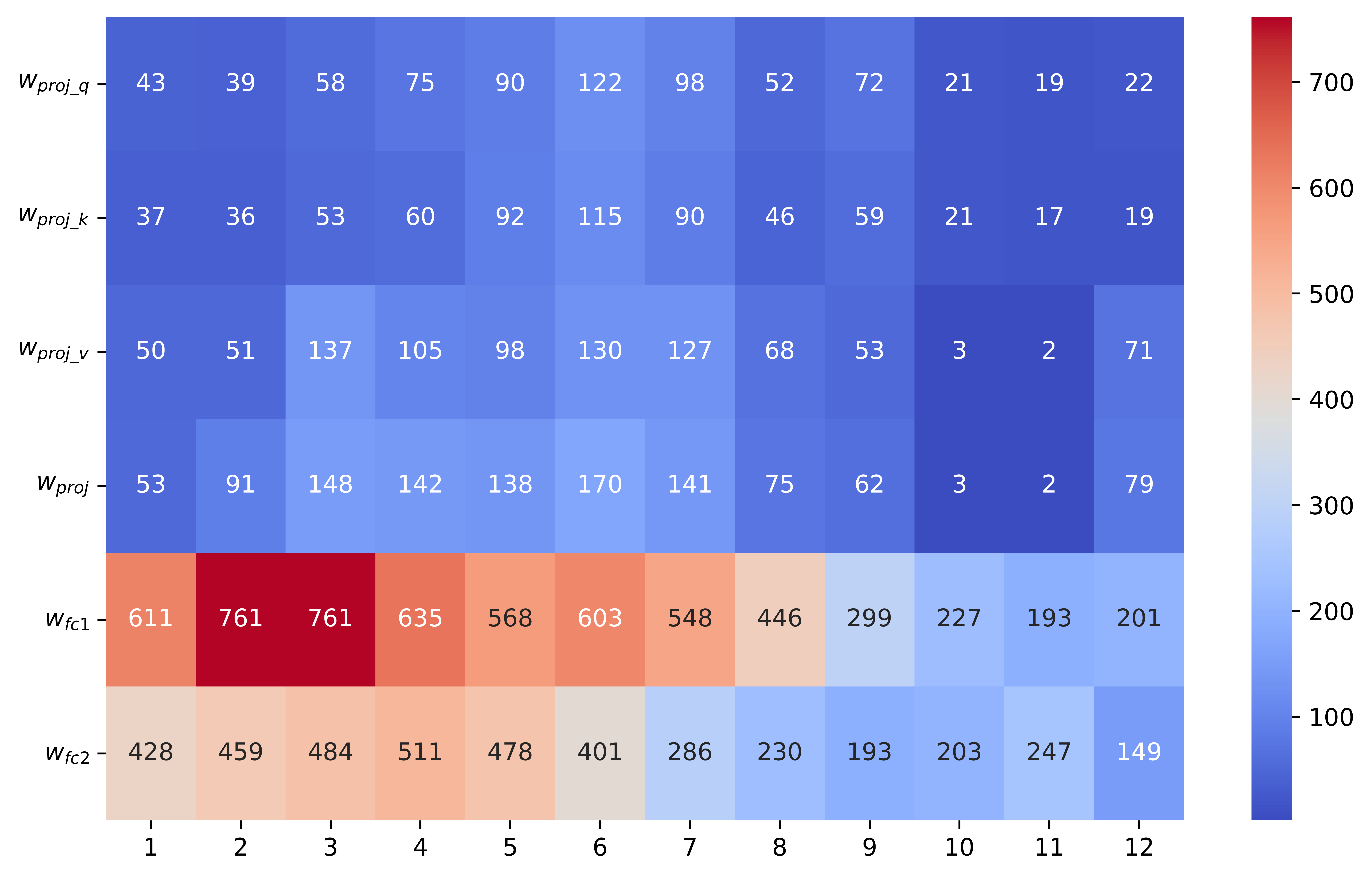}
  \caption{CoLA.}
  \label{fig:cola}
\end{figure}

\begin{figure}[t]
  \centering
  \includegraphics[width=\columnwidth]{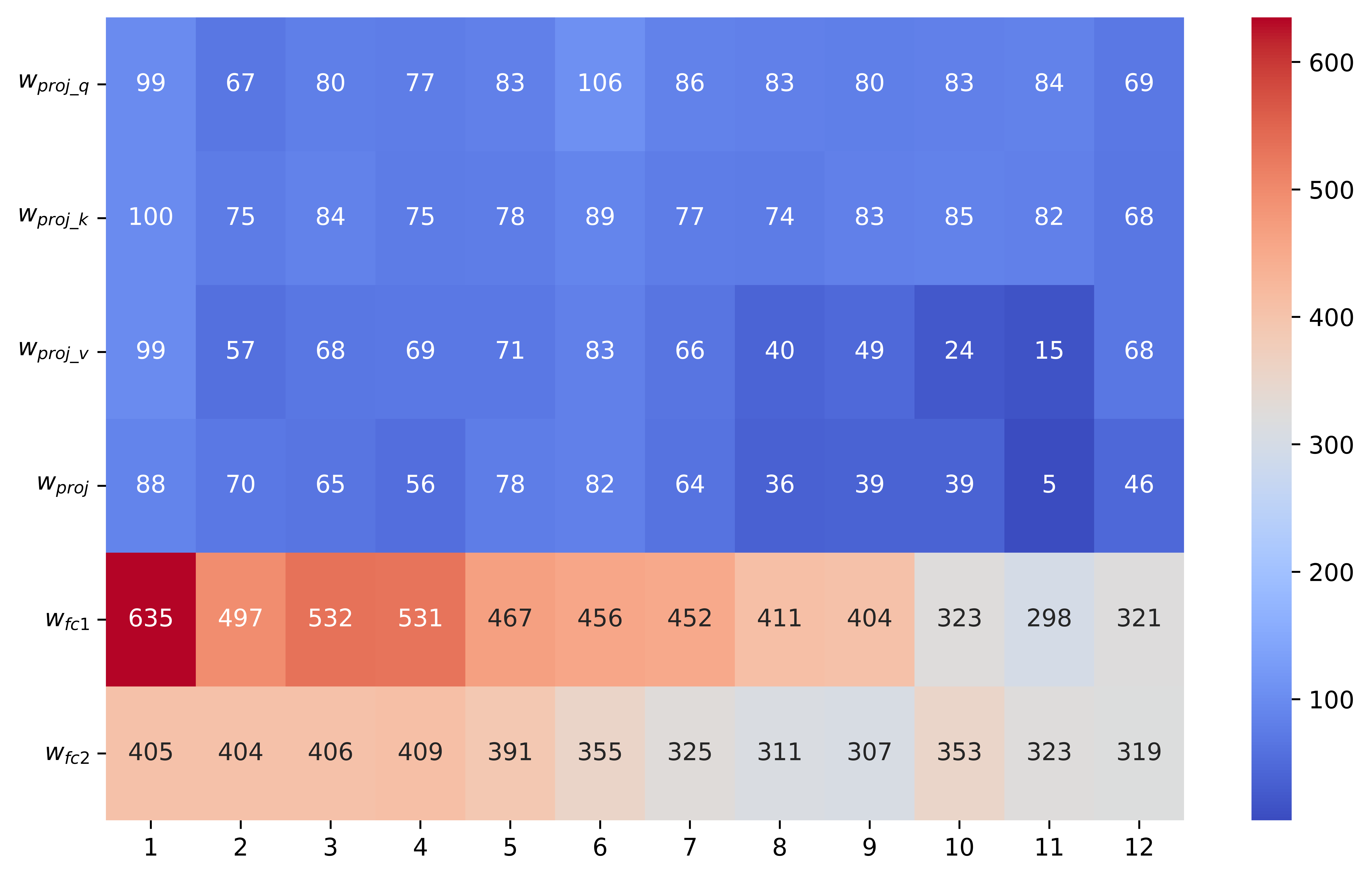}
  \caption{SST-2.}
  \label{fig:sst-2}
\end{figure}

\begin{figure}[t]
  \centering
  \includegraphics[width=\columnwidth]{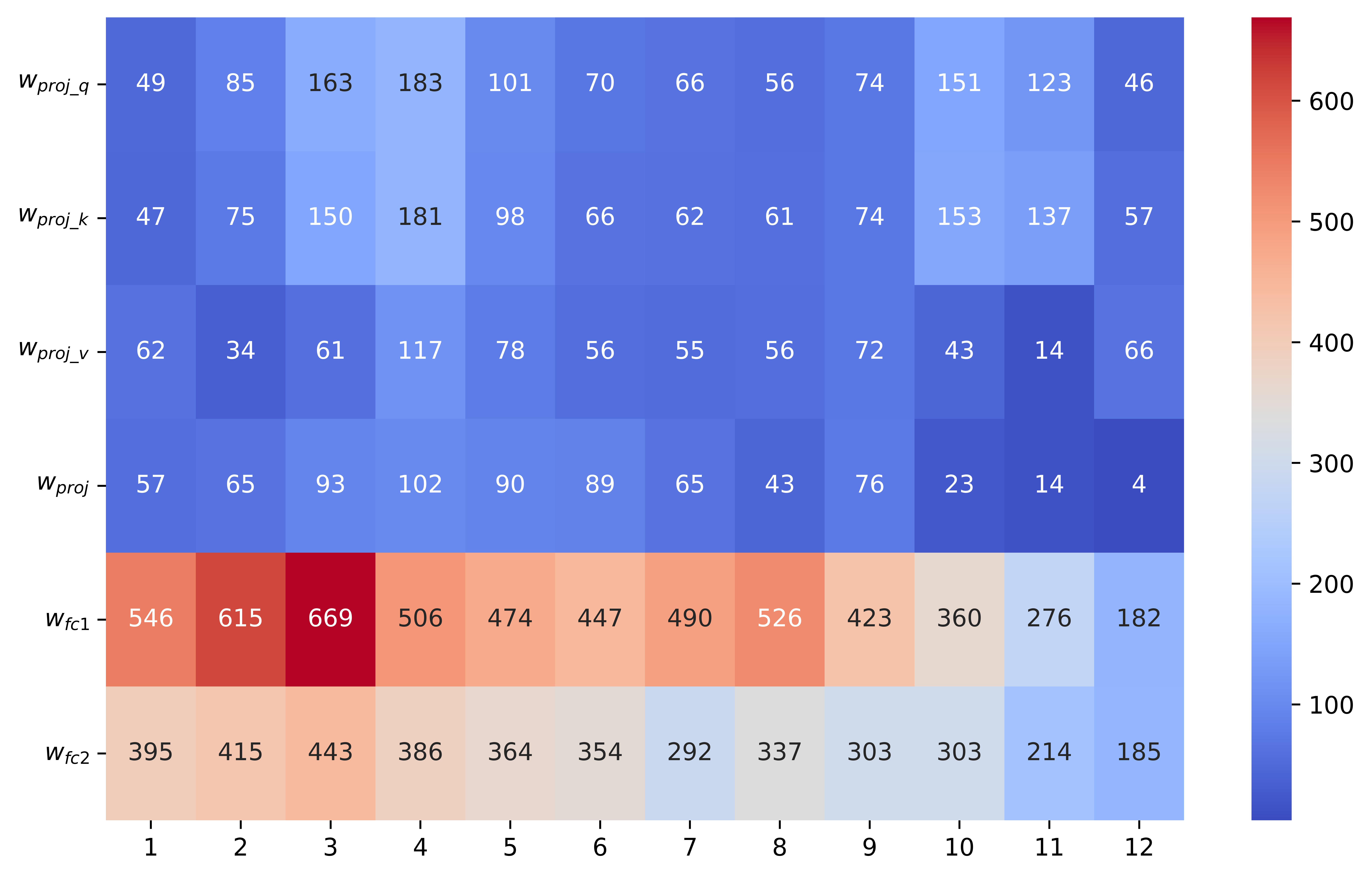}
  \caption{QNLI.}
  \label{fig:qnli}
\end{figure}

\begin{figure}
  \centering
  \includegraphics[width=\columnwidth]{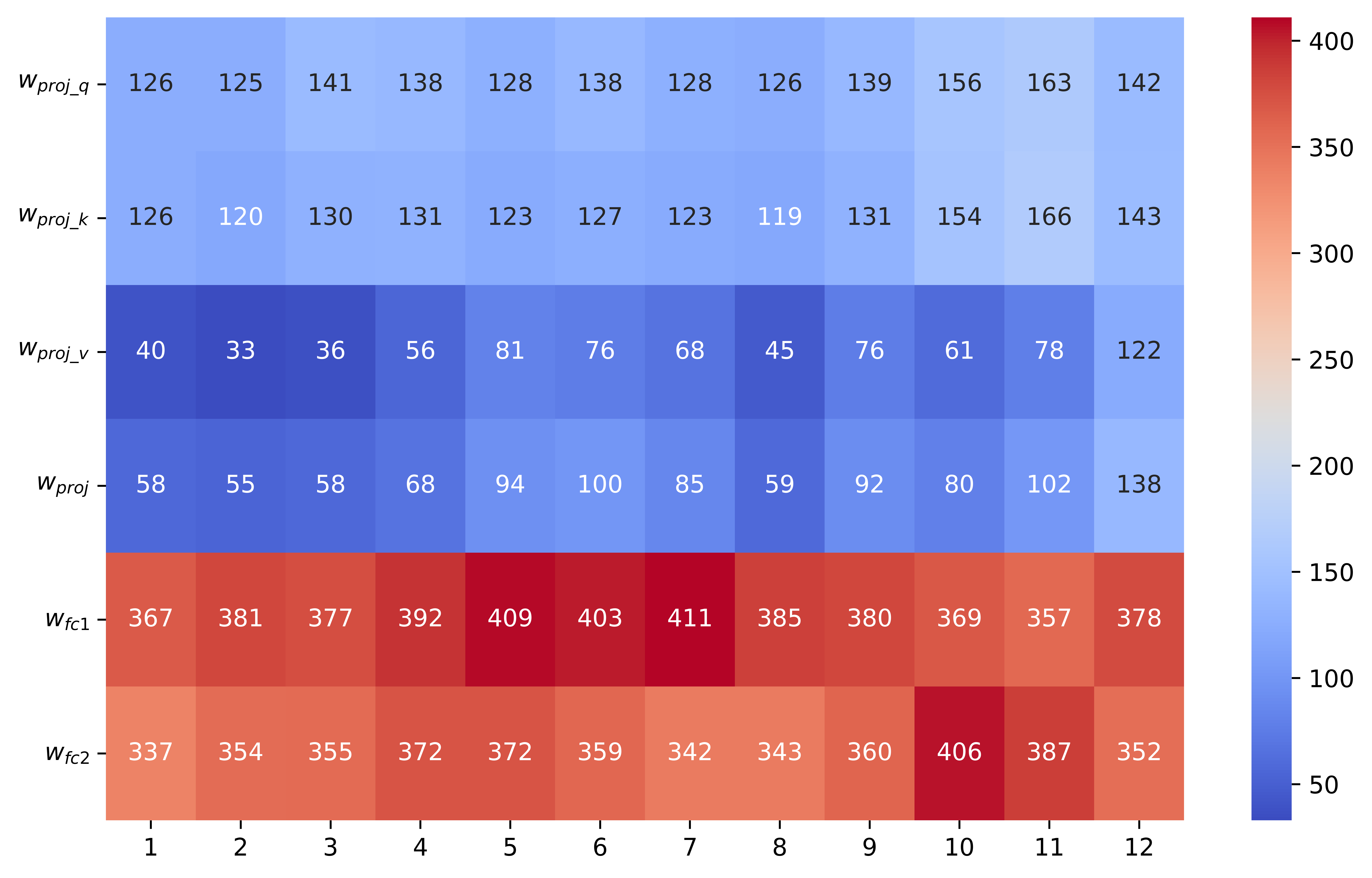}
  \caption{QQP.}
  \label{fig:qqp}
\end{figure}

SoftBERT has the following rank distribution for the six linear layers -  $W_{Q}$, $W_{K}$, $W_{V}$, and $W_{proj}$ in the attention module, $W_{fc1}$ and $W_{fc2}$ in the FFN as illustrated in  Figures~\ref{fig:mrpc} - \ref{fig:qqp} where $x$-axis is the block number. These results indicate that the initial layers have higher importance than later ones. The layers in the FFN have  four times higher parameter counts than the ones in the attention leading to higher ranks. It is also noted that the $W_{fc2}$ achieves better compression than $W_{fc1}$.

\end{document}